%% file: root.tex
\documentclass[letterpaper, 10 pt, conference]{ieeeconf}  

\IEEEoverridecommandlockouts                              

\usepackage{graphics} 
\usepackage{epsfig} 
\usepackage{mathptmx} 
\usepackage{amsmath} 
\usepackage{amssymb}  
\usepackage{booktabs}
\usepackage{makecell}
\usepackage{float}  
\usepackage{subfigure}  
\usepackage{multicol}
\usepackage{wrapfig}
\usepackage{url}
\usepackage{color}
\usepackage{tabularx}
\usepackage{algorithm}
\usepackage[noend]{algorithmic}
\usepackage{capt-of,etoolbox}
\usepackage[export]{adjustbox}
\usepackage{arydshln}
\usepackage[titlenumbered,ruled,noend,algo2e]{algorithm2e}
\usepackage{caption}
\usepackage{multirow}
\usepackage{placeins}
\usepackage{fancyhdr}

\newcommand{\B}{\mathcal{B}}

\newcommand{\numDims}{d_{\mathcal{A}}}

\newcommand{\wx}[1]{\textcolor{blue}{\textbf{#1}}}

\title{\LARGE \bf
GeRM: A Generalist Robotic Model with Mixture-of-experts for Quadruped Robot
}

\author{
Wenxuan Song,
Han Zhao,
Pengxiang Ding,
Can Cui,
Shangke Lyu,
Yaning Fan,
Donglin Wang* \\
     \normalsize MiLAB, Westlake University, China\\
     }

\begin{document}

\twocolumn[{%
\renewcommand\twocolumn[1][]{#1}%
\maketitle
\begin{center}
    \vspace{-0.2in}
    \centering
    \captionsetup{type=figure}
    \includegraphics[width=0.99\linewidth]{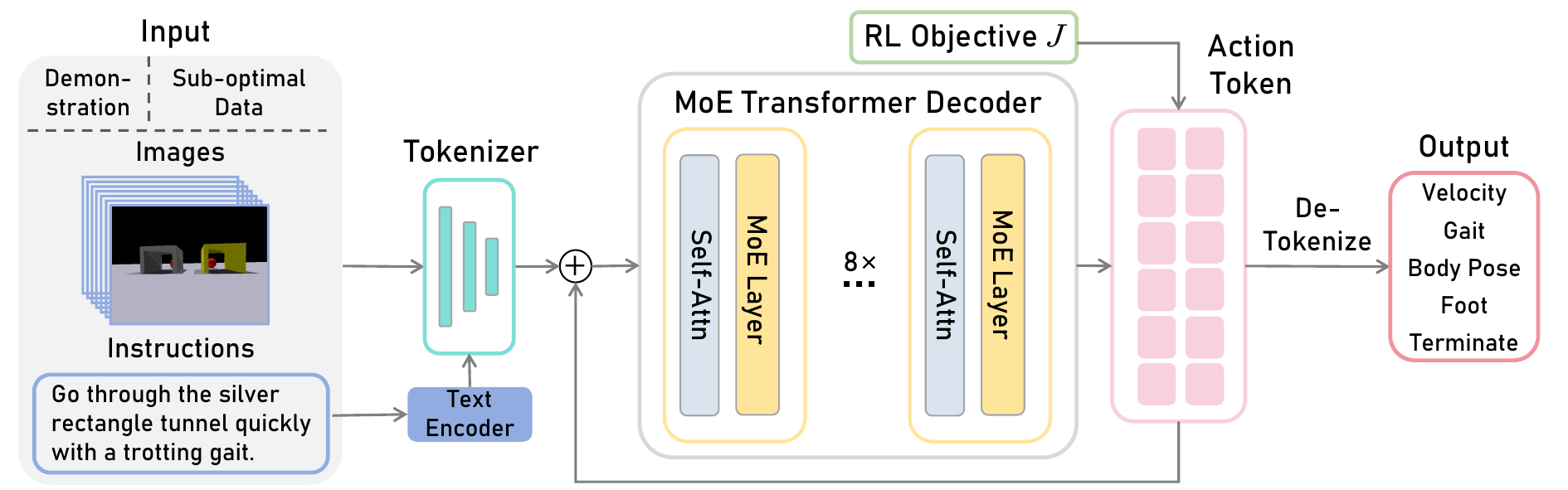}
    \vspace{-0.05in}
    \captionof{figure}{{\small{\textbf{Overview of GeRM.} We take both demonstration and sub-optimal data as input. Then the images and instructions are tokenized and sent into the mixture-of-experts Transformer Decoder to generate action tokens. They are finally de-tokenized into discretized robot commands. The actions are used for RL objectives when training.}}}
    \label{fig:teaser}
    \vspace{-0.05in}
\end{center}
}]

\input{paper/abstract}
\input{paper/intro}
\input{paper/Related_Work}
\input{paper/Preliminaries}
\input{paper/Method}

\input{paper/Experiment}
\input{paper/Conclusion}



\bibliographystyle{./IEEEtran} 
\bibliography{00BIB}


\end{document}

%% file: paper/abstract.tex
\begin{abstract}
Multi-task robot learning holds significant importance in tackling diverse and complex scenarios. 
However, current approaches are hindered by performance issues and difficulties in collecting training datasets. 
In this paper, we propose GeRM (\textbf{Ge}neralist \textbf{R}obotic \textbf{M}odel).
We utilize offline reinforcement learning to optimize data utilization strategies to learn from both demonstrations and sub-optimal data, thus surpassing the limitations of human demonstrations. 
Thereafter, we employ a transformer-based VLA network to process multi-modal inputs and output actions. 
By introducing the Mixture-of-Experts structure, GeRM allows faster inference speed with higher whole model capacity, and thus resolves the issue of limited RL parameters, enhancing model performance in multi-task learning while controlling computational costs. 
Through a series of experiments, we demonstrate that GeRM outperforms other methods across all tasks, while also validating its efficiency in both training and inference processes. Additionally, we uncover its potential to acquire emergent skills.
Additionally, we contribute the QUARD-Auto dataset, collected automatically to support our training approach and foster advancements in multi-task quadruped robot learning. 
This work presents a new paradigm for reducing the cost of collecting robot data and driving progress in the multi-task learning community.

You can reach our project and video through the link: https://songwxuan.github.io/GeRM/ .
\end{abstract}

%% file: paper/intro.tex
\section{Introduction}

\begin{figure*}[t]
    \centering
    \captionsetup{type=figure}
    \includegraphics[width=0.99\linewidth]{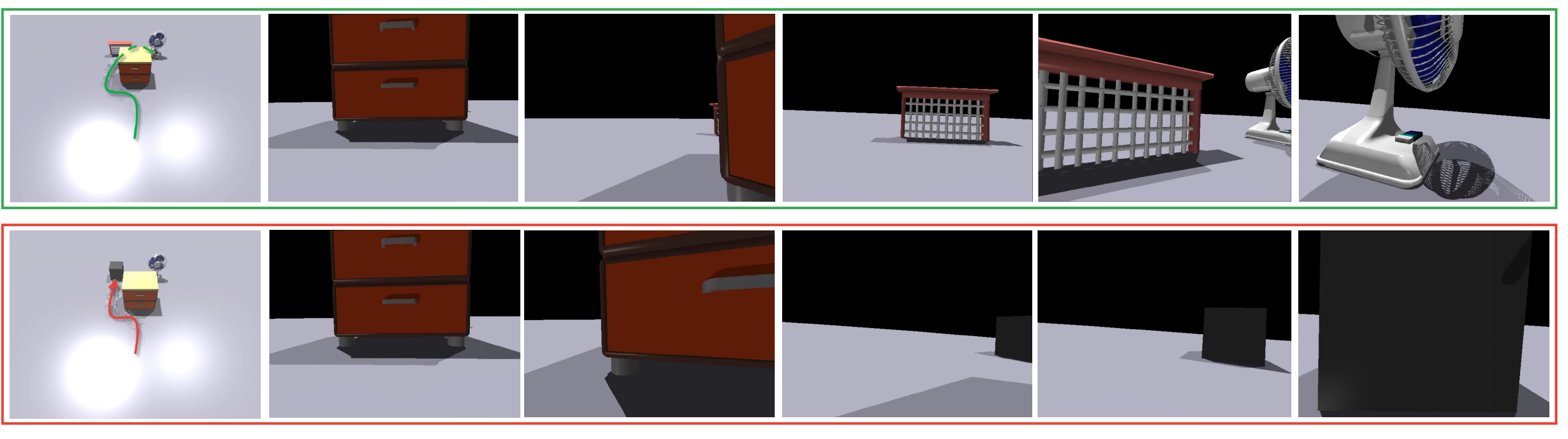}
    \vspace{-0.05in}
    \captionof{figure}{{\small{\textbf{Emergent Skills.} 
   The example of the emergent skill of dynamic adaptive path planning.
   We study these challenging scenarios in detail in Section~\ref{Q4}}}.}
    \label{fig:emergent}
    \vspace{-0.2in}
\end{figure*}

Quadruped robots, known for their exceptional ability to traverse complex terrains and execute agile movements, have become a focal point in robotics research~\cite{Hutter2016anymal,lyu2023composite}. 
Researchers have extensively utilized these robots to tackle various tasks, including autonomous navigation (e.g. urban navigation \cite{vinl, karnan2022scand}), locomotion \cite{Lee2020LearningQL, scirobotics, Yang2021LearningVQ}, manipulation \cite{wholebody}, and also multi-task learning \cite{mtopt, Pre-Training_2022}.

To achieve the capability to handle multi-task scenarios, quadruped robots should have the ability to receive human instructions, perceive the environment, autonomously make plans, and take action. 
Therefore, we want to combine language and visual inputs and output actions by utilizing the Vision-Language-Action (VLA) model proposed in RT-1 ~\cite{brohan2023rt1} into quadruped robot learning.

%

However, the existing VLA models, which rely on expert data collected for \textbf{Imitation Learning (IL)}, have the following problems:

\wx{1.} The cost of manually collecting datasets is high. 
IL training relies on large-scale robot datasets~\cite{openx}. 
Current methods for collecting robot data are based on real-world environment \cite{ebert2021bridge,walke2023bridgedata}, which requires experts' remote control, and simulation environment \cite{Mees2021CALVINAB}, which requires environment setup and algorithm design.
Meanwhile, as the robot with the most degrees of freedom (DOFs), the difficulty in controlling quadruped robots is also notably high.
These factors contribute to the increased difficulty and cost associated with collecting high-quality expert quadruped data.
Therefore, we hope to automatically collect datasets and utilize them for training.

\wx{2.} The performance of the IL policy is limited by the degree to which experts can provide high-quality demonstrations.
This paper aims to employ \textbf{Reinforcement Learning (RL)} methods to learn from auto-collected datasets and reasonably utilize sub-optimal data to break through the demonstration. 
To utilize pre-collected large-scaled datasets, we choose the offline RL algorithm.
Then the core issue is how to effectively apply the transformer-based VLA model to offline RL.
Effective offline RL generally employs Deep Q-Learning.
Therefore, we adopt designs akin to Q-Transformer \cite{qtransformer} by employing a transformer-based VLA model to replace the value function and output discretized actions.

The augmentation of parameter quantity frequently enhances a model's capacity for generalization across multi-tasks, which has been proved in many fields \cite{brown2020gpt3,brohan2023rt2}.
However, augmenting the parameter count of an RL policy often negatively impacts its overall performance.
Recently, \cite{obando2024mixtures} has proved the effectiveness of mixture-of-experts (MoE) to unlock parameter scaling in deep RL.
Thus, we construct a mixture-of-experts structure. 


GeRM is a sparse MoE network \cite{jacobsmoe, jordanmoe}. 
It is a transformer decoder-only model where the Feed-Forward Network (FFN) picks from a set of \textbf{8} distinct groups of parameters.  
At every layer, for every token, a router network chooses two of these groups (the “experts”) to process the token and combine their output additively. 
Different experts are proficient in different tasks/different action dimensions to solve problems in different scenarios, learning a generalist model across multiple tasks. 
This technique increases the network parameter volume while keeping the computational cost basically unchanged, as the model only uses a fraction of the total set of parameters per token.

We collected the QUARD-Auto dataset in an automatic collection manner as a supplement to our previously published QUARD dataset~\cite{Ding2023QUARVLAVM}, addressing the shortcomings of failed (sub-optimal) data. 
It must be emphasized that we have explored a fully automated approach to data collection, which circumvents the difficulties and costs associated with manually controlling robots for demonstrations. 
We simply provide instructions and utilize the pre-trained VLA model to autonomously control the robot, thereafter recording both the received image and the executable action, resulting in the collection of \textbf{258418} trajectories on Issac Gym, comprising \textbf{120128} success and \textbf{138290} failures.
This presents a new paradigm for the autonomous collection of large-scale robot datasets.

Our contributions mainly lie in two aspects:
\begin{itemize}
    \item We first propose a Mixture-of-Experts model for quadruped reinforcement learning.
    We have adopted a Mixture-of-Experts structure to replace the conventional linear layer within the Transformer decoder, which allows faster inference speed with higher whole model capacity. Additionally, deep Q-learning methodology aims to acquire and optimize the model's capabilities to its optimal potential.
    \item  We have extensively validated the effectiveness of GeRM through numerous experiments. 
    It has been trained on limited demonstrations and sub-optimal data, then extensively tested across \textbf{99} tasks. 
    GeRM outperforms existing methods and exhibits superior capabilities across multi-tasks, with only \textbf{1/2} total parameters activated. 
    Furthermore, other experiments also demonstrate GeRM's superiority in data utilization and emergent skill development.
    \item We contributed an auto-collected dataset with failed data that can be used for reinforcement learning, enabling learning on sub-optimal data, thus breaking through the limitations of human demonstration data. 

\end{itemize}

%% file: paper/Related_Work.tex
\section{Related Work}
\input{tabletex/tab1_tasks}
\textbf{Offline RL for Legged Robot Control.}
Recent works have extensively explored offline RL. \cite{Jaques_Ghandeharioun_Shen_Ferguson_Lapedriza_Jones_Gu_Picard_2019,Wu_Tucker_Nachum_2019,Peng_Kumar_Zhang_Levine_2021,Siegel_Springenberg_Berkenkamp_Abdolmaleki_Neunert_Lampe_Hafner_Heess_Riedmiller_2020,Kostrikov_Nair_Levine_2021,Fujimoto_Gu_2021,Chen_Zhou_Wang_Che_Wu_Ross_2019,Furuta_Matsuo_Gu,Jang_Lee_Kim,Meng_Wen_Le_Li_Xing_Zhang_Wen_Zhang_Wang_Yang_et,Liu_Tang_Li_Luo}, with Conservative Q-learning (CQL) \cite{Kumar_Zhou_Tucker_Levine_2020} focusing on learning policies that adhere to a conservative lower bound of the value function. The objective of our research is to create an offline RL framework capable of seamless integration with high-capacity Transformers and scalable for multi-task robotic learning. 
Q-Transformer \cite{qtransformer} developed a variant of CQL specifically optimized for training large Transformer-based Q-functions on mixed-quality data. 
Our work is aimed at training more general and efficient strategies based on this type of framework.

\textbf{Sparse Mixture-of-Experts Architecture.}
Sparse Mixture-of-Experts models have shown significant advantages in natural language processing (NLP). 
\cite{SparseGatedMoE} showed that they could effectively use a very large number of weights while only activating a small subset of the computation graph when inference, which explains the term ``sparse".
There has also been work on scaling sparse MoE architecture\cite{Hestness_Narang_Ardalani_Diamos_Jun_Kianinejad_Patwary_Yang_Zhou_2017} and apply it on Transformers\cite{GShard} \cite{Kudugunta_Huang_Bapna_Krikun_Lepikhin_Luong_Firat_2021} \cite{effectivemoe}.
Within it, \cite{Switch_Transformer_2021} and \cite{du2022glam} have expanded the MoE model capacity to 1 trillion parameters.
Recently, in the era of LLM, MoE has become a broad and effective structure \cite{jiang2024mixtral} \cite{dai2024deepseekmoe}.
MoE has also helped deep RL with parameter scalability \cite{obando2024mixtures}.
Now we aim to apply MoE on robotic control to obtain a generalist model.

\textbf{Transformer-based Vision-Language-Action Model.}
VLA models (\cite{Shridhar2021CLIPortWA, reed2022gato, Nair2022R3MAU,brohan2023rt1,brohan2023rt2, bharadhwaj2023roboagent,Li2023roboflamingo,szot2023llarp}) integrates visual information and instructions to generate executable actions.
Transformer-based VLA models hold the potential to handle general tasks by processing general inputs and outputs.  
Our previous work \cite{Ding2023QUARVLAVM} has pioneered the deployment of the VLA model on quadruped robots. 
While existing VLA models are typically trained using imitation learning approaches, Q-Transformer \cite{qtransformer} was the first to employ RL methods for training VLA models.
We intend to further enhance the training of VLA models for quadruped robots using RL in a more effective manner.

%% file: tabletex/tab1_tasks.tex
\begin{table*}[t]
\caption{
\textbf{Illustration of tasks.}  The ``Skill" means different skill/task categories. 
The ``Episode" signifies the number of experiments conducted for each task, which also corresponds to the number of trajectories.  
The ``Description" is the description of the tasks.
The ``Example Instruction" describes different task scenarios, including various higher-level variables associated with the simulation.  
}
\label{tab:seentasks}
\renewcommand\arraystretch{1.1}
\scriptsize
\centering
\setlength{\tabcolsep}{0.78mm}{
\begin{tabular}{p{0.22\columnwidth}p{0.15\columnwidth}p{0.7\columnwidth}p{0.7\columnwidth}} 
\hline
\textbf{Skill}  & \textbf{Episode} & \textbf{Description} & \textbf{Example Instruction} \\
\hline
Go to \textit{Object} & 66K & Navigate to the object and stop in front of it & Go to the trashcan slowly with a trotting gait. \\
Go to \textit{Object} and avoid the obstacle  & 47K & Navigate to the object without colliding with the obstacle & Go to the piano and avoid the obstacle quickly with a bounding gait. \\
Stop \textit{Object}  & 51K & Move to block the ball rolling toward the robot & Stop the red ball normally with a pacing gait. \\
Distinguish \textit{Letter} & 16K & Identify the correct one from multiple boxes with different printed letters & Distinguish letter B normally with a bounding gait. \\
Go through \textit{Tunnel}  & 77K & Go through the correct tunnel from two tunnels with different colors and shapes & Go through the silver rectangle tunnel quickly with a trotting gait. \\

\hline
Total    & 257K & The total number of episodes & \\

\hline

\end{tabular}
}
\vspace{-1.em}
\end{table*}

%% file: paper/Preliminaries.tex
\section{Preliminaries}
\label{sec:preliminary}
In RL, for a Markov decision process (MDP), there is a state $s$, actions $a$, discount factor $\gamma \in (0, 1]$, transition function $T(s'|s,a)$ and a reward function $R(s,a)$. 
In RL, we learn policy $\pi$ that maximizes the expected total reward in a Markov decision process (MDP) with states $s$, actions $a$, discount factor $\gamma \in (0, 1]$, transition function $T(s'|s,a)$ and a reward function $R(s,a)$. Actions $a$ have dimensionality $\numDims$. 
Value-based RL approaches learn a Q-function $Q(s,a)$ representing the total discounted return $\sum_t \gamma^t R(s_t, a_t)$, with policy $\pi (a|s) = {\operatorname{argmax}_{a}} Q(s,a)$.
The Q-function can be learned by iteratively applying the Bellman operator:
\begin{align}
\B^* Q(s_t,a_t) = R(s_t,a_t) + \gamma \max_{a_{t+1}}Q(s_{t+1}, a_{t+1}),
\vspace{-3pt}
\end{align}
approximated via function approximation and sampling. 

Then, following the setting in Q-Transformer, we need to apply discretization and autoregression by regarding each action as a different dimension:




\begin{align}
    Q(s_{t-w:t}, a_t^{1:i-1}, a_t^i) &= \notag\\
    &\begin{aligned}[t]
    \hspace{-8em}\begin{cases} 
     \underset{a_t^{i+1}}{\max} \mkern9mu Q(s_{t-w:t},  a_t^{1:i}, a_t^{i+1}) & \text{if } i \in \{1, \dots, d_\mathcal{A}-1\}\\
     R(s_t, a_t) + \gamma \underset{a_{t+1}^1}{\max} \mkern9mu Q(s_{t-w+1:t+1}, a_{t+1}^1) & \text{if } i = d_\mathcal{A} 
    \end{cases}
    \end{aligned}\label{eq:q_update}
\end{align}
where $\tau = (s_1, a_1, \dots, s_{T}, a_{T})$ is a trajectory of robotic experience of length $T$ from an offline dataset $\mathcal{D}$.  $t$ is a given time-step, and $a_t$ is the corresponding action in the trajectory, $a^{1:i}_t$ denote the vector of action dimensions from the first dimension $a^1_t$ until the $i$-th dimension $a^i_t$, $i$ can range from $1$ to the total number of action dimensions $\numDims$, $w$ is a time window of state history.

To tackle the out-of-distribution question in offline datasets, we add a conservative penalty~\cite{Kumar_Zhou_Tucker_Levine_2020} that pushes down the Q-values $Q(s,a)$ for any action $a$ outside of the dataset, thus ensuring that the maximum value action is in-distribution.
In CQL, let $\pi_\beta$ be the behavioral policy that induced a given dataset $\mathcal{D}$, and let $\tilde{\pi}_\beta$ be the evaluation policy.
Our objective to train the Q-function is: 
\begin{align}
    \hspace{-2pt} J &= ~\frac{1}{2}~ \mathbb{E}_{s \sim \mathcal{D},a \sim \pi_\beta(a|s)} \left[\left(Q(s,a) - \mathcal{B}^* Q^{k}(s,a)\right)^2\right] \notag \\
    &\quad + \alpha \cdot \frac{1}{2}\mathbb{E}_{s \sim \mathcal{D},a \sim \tilde{\pi}_\beta(a|s)} \left[(Q(s,a) - 0) ^2\right], \label{eq:opt_obj}
\end{align}
where the first term trains the Q-function by minimizing the temporal difference error objective as defined in Eq.~\ref{eq:q_update}, and the second term regularizes the Q-values to the minimal possible Q-value of $0$ in expectation under the distribution of actions induced by $\tilde{\pi}_\beta$, which we denote as a conservative regularization term $\mathcal{L}_C$, $\alpha$ is a factor which modulates the strength of the conservative regularization. 

%% file: paper/Method.tex
\section{Methods}

\subsection{Auto-collected Quadruped Robot Datasets}
To effectively train a generalist model through RL, it is essential to facilitate the seamless collection of a diverse dataset, including successful data and failed data, enabling corrective feedback and scalable task evaluation 
Therefore, we collect a large-scale multi-task dataset, \textbf{QUARD-Auto}, which includes multiple tasks such as navigation and whole-body manipulation.
Next, we will discuss the main components of our data collection process.

\begin{figure}[t]
  \centering
  \includegraphics[width=0.5\textwidth]{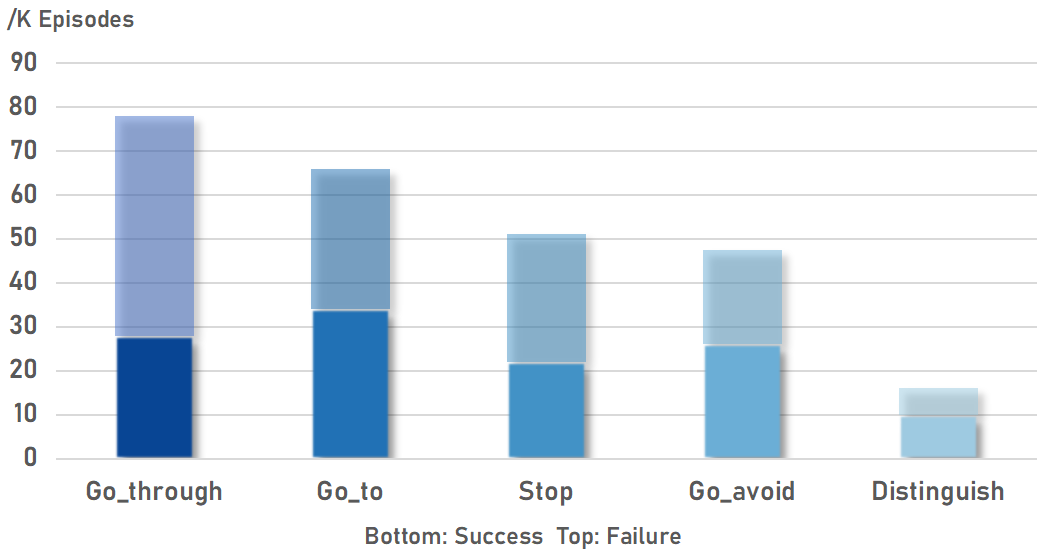}
  \caption{\textbf{Statistic of QUARD-Auto.} The \textbf{Bottom} parts denote the successful tasks; the \textbf{Top} parts denote the failed tasks.}
  \label{fig:dataset}
\end{figure}

\textbf{Environment and Tasks.}
In this paper, we define and collect the data of {5} kinds of tasks. 
The detailed list of tasks in the training dataset is shown in Table~\ref{tab:seentasks}. 
The data was collected in Nvidia's Isaac Gym~\cite{makoviychuk2021isaac}, a powerful simulator that allows us to collect massive robot trajectories in parallel.
More statistical details about QUARD-Auto can be seen in Figure~\ref{fig:dataset}.
Different tasks correspond to different success criteria.
For example, in the ``\textit{Go to}", ``\textit{Go avoid}", and ``\textit{Go through}" tasks, the success condition is to reach a specified location. 
The success condition for ``\textit{Stop}" is to touch and stop the moving object and the success condition for ``\textit{Distinguish}" is to turn to the selected visual target.

\textbf{Data Collection.}
For simulated data collection, the robot uses a combination of low-level and high-level control. 
The high-level control combines path planning with robot locomotion according to the global spatial information of the robot, obstacles, and target objects. 
For autonomous collection, we directly utilize a pre-trained policy to eliminate any need for manual teleoperation or specific trajectory design.
Here, we utilize GeRM w/o MoE pre-trained on demonstrations as our high-level policy, which can receive instructions (from a simple pre-written template) and images (from a camera in the simulated environment) and output commands, eventually forming complete trajectories.
The low-level control deploys the command data output by the high-level policy into actual robot actions.
Here, we adopt the approach proposed in \cite{margolis2023walk} as the pre-trained low-level control strategy to output actual robot joint angles.
We collected instructions, images, and command data for each frame and ultimately obtained a mix of successful and unsuccessful data.


\subsection{Mixture-of-Experts Network}
\label{subsec:mixture-of-experts-network}

\begin{table}[t]
\small
\centering
\begin{tabular}{lr}
\toprule
\textbf{Parameter}  & \textbf{Value} \\ \midrule
\texttt{action\_dim}             & $12$           \\
\texttt{num\_layers}       & $8$             \\
\texttt{layer\_size}       & $4096$             \\
\texttt{num\_heads}        & $8$             \\
\texttt{num\_kv\_heads}    & $8$              \\
\texttt{context\_len} & $512$           \\
\texttt{time\_length} & $7$           \\
\texttt{vocab\_size}     & $256$          \\ 
\texttt{num\_experts}  & $8$ \\
\texttt{top\_k\_experts}  & $2$ \\
\bottomrule
\end{tabular}
\caption{\small \textbf{Model architecture.}}
\label{tab:param}
\end{table}

GeRM is based on a transformer architecture \cite{vaswani2017attention} and consists of 8 self-attention layers and 167M total parameters that outputs action tokens, and the FFNs are replaced by MoE layers.
The model architecture parameters are summarized in Table~\ref{tab:param}.

\begin{figure}[t]
  \centering
  \includegraphics[width=0.5\textwidth]{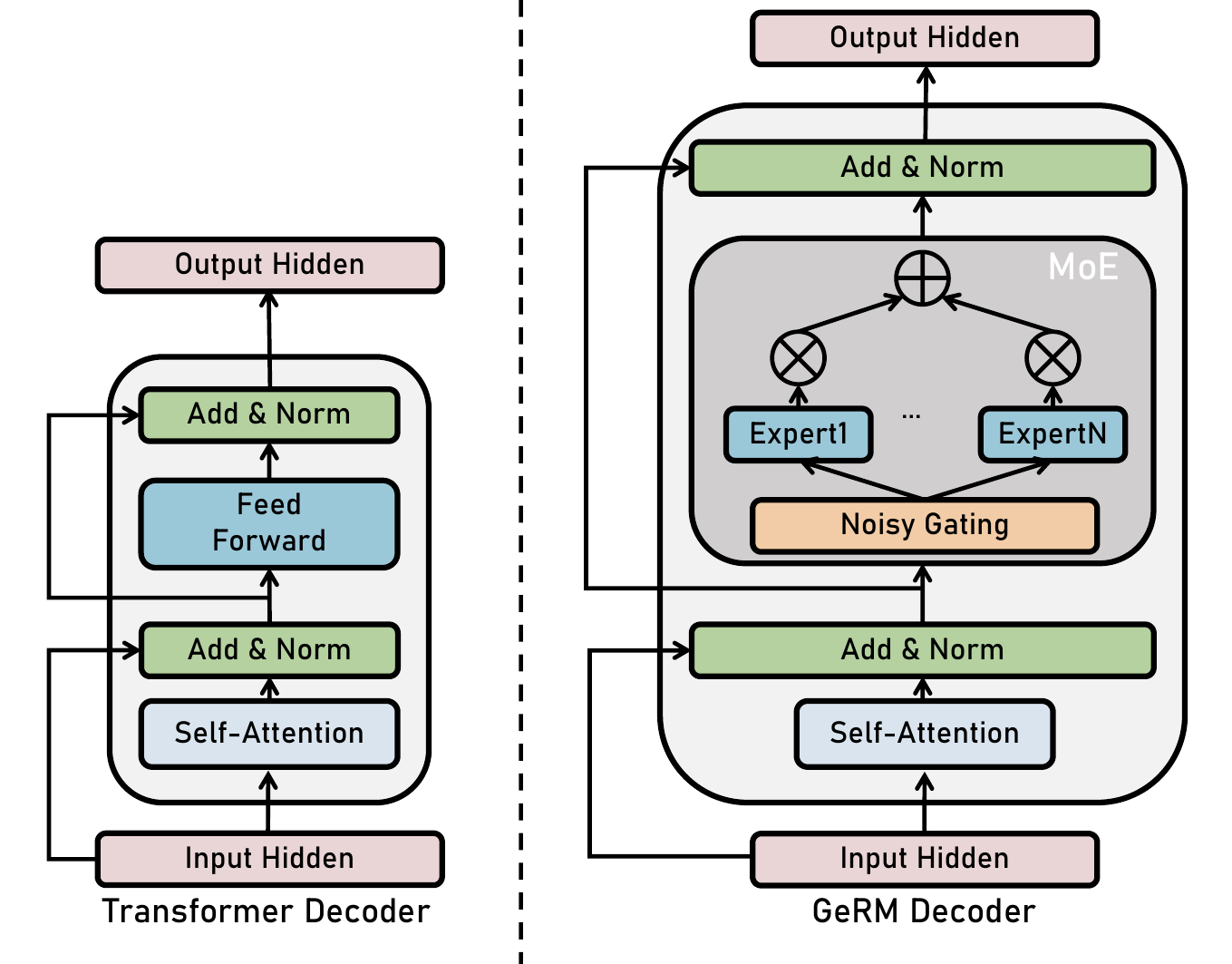}
  \caption{\textbf{Decoder Structure.} \textbf{Left:} Conventional Transformer Decoder;~\textbf{Right:} GeRM Transformer Decoder with MoE Module.}
  \label{fig:MoE}
  \vspace{-0.2in}
\end{figure}

We present a brief overview of the Mixture-of-Experts layer in Figure~\ref{fig:MoE}. 
The MoE module's output for a given input $x$ is computed through the weighted sum of the expert networks' outputs, where the weights are given by the gating networks $G$. 
Then the output $y$ could be described as:
\begin{equation}
   y = \sum_{i=0}^{n-1} G(x)_i \cdot E_i(x),
\end{equation}
where $n$ is the number of expert network, the $G(x)_i$ denotes the $n$-dimensional output of the gating network for the $i$-th expert, and $E_i(x)$ is the output of the $i$-th expert network.
There are multiple alternative ways of implementing $G$~\cite{zhou2022mixture},~\cite{obando2024mixtures}, and one simple but effective way is implemented by taking the softmax over the Top-K logits of a linear layer.
Before taking the softmax function, we add tunable Gaussian noise, which helps with load balancing - the Gaussian noise term adds randomness while making the process of obtaining discrete quantities from continuous quantities differentiable, thereby allowing for the back-propagation of gradients.
We use
\begin{equation}
\begin{aligned}
G(x) & = \text{Softmax}(K(H(x), k)) \\
     & = \frac{\exp(k(x)_i)}{\sum_{j=0}^{N-1} \exp(k(x)_j)} \quad \\
     & \text{for} \quad i = 0, 1, 2, ..., n-1 ,
\end{aligned}
\end{equation}
$H(x)$ is implemented by
\begin{equation}
\begin{aligned}
H(x)_i = (x \cdot W_g)i + \mathcal{N}(0, 1) \cdot \text{Softplus}((x \cdot W_{\text{noise}})_i),
\end{aligned}
\end{equation}
where $W_g$ denotes the weights of gates, and $K(x,k)$ is implemented by 
\begin{equation}
\begin{aligned}
K(x, k) & = \text{TopK}(x \cdot W_g) \\
     & = \begin{cases}
            x \cdot W_g, & \text{if } x \text{ is in the } \text{TopK} \text{ elements.} \\
            -\infty, & \text{otherwise.}
         \end{cases}
\end{aligned}
\end{equation}
where $k$ in TopK denotes the number of experts used per token, it is a hyperparameter that modulates the amount of compute used to process each token. When $n$ is changed while K is fixed, the model's parameters could be changed while its computational cost is still constant. 
Therefore we also called the model's total parameter count the sparse parameter count and the parameters for processing an individual token the active parameter count, which means parameters actually used when inference.

\subsection{Vision-Language-Action Model in Reinforcement Learning}
An overview of GeRM is shown in Figure~\ref{fig:teaser}.
In GeRM, the instruction is first processed via universal sentence encoder~\cite{Universal_sentence_encoder}~$E_{text}(z_i|s)$ to get 512-dimension vectors~$z_i$, then sent into the ImageNet-pretrained EfficientNet-B3~\cite{jiang2011efficient} with FiLM~\cite{Perez2017FiLMVR}~$q_v(z_v|s,z_i)$ together with the history of 6 (the 7th image only for calculating Q-value) images~$w$ to got vision-language tokens~$z_v$.
The resulting vision-language tokens $z_v$ are followed by a TokenLearner~\cite{tokenlearner2021}~$\tau(t|z_v)$ to compute a compact set of tokens~$t$, and finally MoE Transformer decoders~$p_{MoE}(a_d|t)$ described in \ref{subsec:mixture-of-experts-network} to attend over these tokens and produce discretized action tokens~$a_d$.
We follow the RL method described in \ref{sec:preliminary} to renew MoE Transformer decoders.
The policy GeRM could be shown as follows: 
\begin{equation}
\begin{aligned}
 &\operatorname {GeRM}(a_d|s, w)  = p_{MoE}(a_d|t) \tau(t|z_v)  q_v(z_v|w, z_i) E_{text}(z_i|s)\\
\end{aligned}
\end{equation}
where $s,w$ are the input images and language instruction and $q_v$ are the language-image feature encoder, $\tau$ represents the token-learner and $p_{MoE}$ indicates the transformer decoder to output action $a_d$.
Eventually $a_d$ is de-tokenized into \textbf{12}-dimensional commands:
\begin{align}
\left[v_x, v_y, \omega_z, \theta_1, \theta_2, \theta_3, f, h_z, \phi, s_y, h_z^f, T \right]
\end{align}
Here, $v_x$, $v_y$, and $\omega_z$ represent the velocities along the x-axis, y-axis, and z-axis respectively. $\theta_1$, $\theta_2$, and $\theta_3$ indicate the gait pattern, $f$ denotes the frequency, $h_z$ represents the height of the robot, $\phi$ denotes the pitch angle, $s_y$ corresponds to the foot width, $h_z^f$ represents the foot height, and $T$ indicates the termination signal of the action.
\input{tabletex/tab3_Experiment}

%% file: tabletex/tab3_Experiment.tex
\begin{table*}[h]
\setlength{\tabcolsep}{0pt}
\scriptsize
\begin{tabular*}{\textwidth}{@{\extracolsep{\fill}}>{\arraybackslash}p{1.65cm}>{\centering\arraybackslash}p{0.9cm}*{8}{>{\centering\arraybackslash}p{0.95cm}}}
\toprule
\textbf{Model}        & \textbf{Total Params} & \textbf{Active Params} & \textbf{\tiny{Sub-optimal Data}} & \textbf{Go\_to}   & \textbf{Go\_avoid} & \textbf{Stop}  & \textbf{Distinguish}   & \textbf{Go\_through}  \\ \midrule
\textbf{RT-1} & 33.50M                                                               & 33.50M          & N            & 48.67          & 33.50         & 42.5         & 44.33         & 0  \\[5pt]
\textbf{GeRM w/o RL}  & 83.48M                                                               & 39.31M          & N             & 49.37          & 46.37         & 44.88         & 52.00         & 28.44  \\[5pt]
\multirow{2}{*}{\textbf{GeRM w/o MoE}}   & \multirow{2}{*}{33.50M}                                                                & \multirow{2}{*}{33.50M}     & N      & 55.01             & 55.44          & 43.93         & 60.73         & 35.34         \\[5pt]
& & &  Y & 62.43             & 60.89         & 45.67         & 63.55         & 47.79      \\\midrule
  
\multirow{2}{*}{\textbf{GeRM}}   & \multirow{2}{*}{83.48M}                                                                & \multirow{2}{*}{39.31M} & N      & 86.37             & \textbf{87.36}          & 50.31         & 75.50         & 73.66\\[5pt]
& & & Y & \textbf{90.50}             &  85.50         & \textbf{71.00}         & \textbf{82.50}         & \textbf{75.00}      \\ \bottomrule
\end{tabular*}
\vspace{2pt}
\caption{
\small
\textbf{Multi-task performance comparison.} \textbf{GeRM} outperforms other models on most tasks while using approximately the same active parameters. The numbers in the table represent the success rate of tasks (\%) . 
}
\label{tab:results}
\end{table*}

%% file: paper/Experiment.tex
\section{Experiments}
In our experiments, we aim to answer the following questions: 
\textbf{Q1.} How does the effectiveness of GeRM as a generalist model, which learns from a combination of demonstrations and sub-optimal data? 
\textbf{Q2.} How important are the specific designs (MoE module, Q-learning) in GeRM? 
\textbf{Q3.} Does the MoE module leverage its strength in size and efficiency in GeRM?
\textbf{Q4.} How does GeRM demonstrate its advantages in training efficiency and data utilization?
\textbf{Q5.} Can GeRM exhibit emergent skills across different tasks?
\begin{figure}[t]
  \centering
  \includegraphics[width=0.5\textwidth]{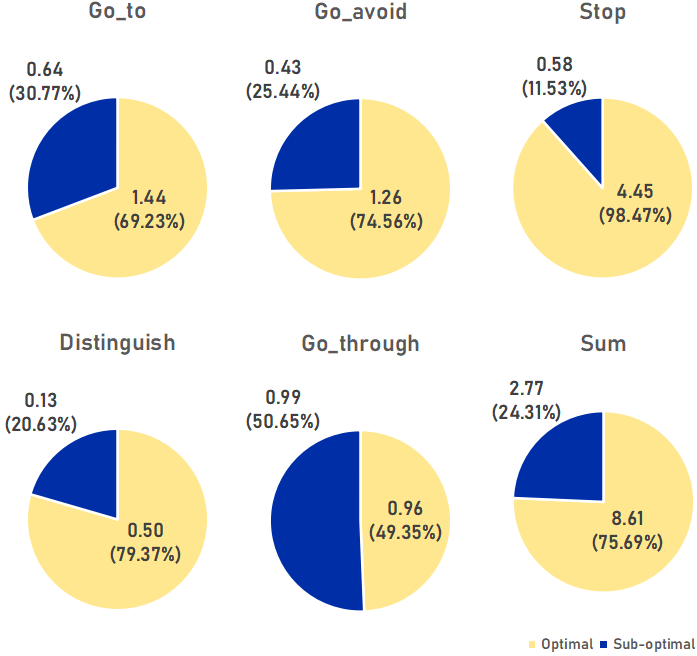}
  \caption{\textbf{Training dataset.} The ratio of the \textbf{optimal trajectories} and \textbf{sub-optimal trajectories} used in training.The unit of trajectory number in the graph is K=10$^3$.}
  \label{fig:train}
  \vspace{-0.2in}
\end{figure}
~\subsection{Experiments Setup}
\textbf{Offline Training Datasets.}
The offline dataset used in our experiment includes 2 categories: demonstrations and sub-optimal data.
Demonstrations correspond to successful tasks, which consist of 5 types of tasks, 99 sub-tasks, with a total of \textbf{8610} trajectories and \textbf{2238600} vision-language-action sets, the length of each trajectory is 260 frames, all sourced from human demonstration data in QUARD \cite{Ding2023QUARVLAVM}.
Sub-optimal data represents failed tasks, which consist of 5 types of tasks,  99 sub-tasks, with a total of \textbf{2766} trajectories and \textbf{1548960} vision-language-action sets, the length of each trajectory is 560 frames, all sourced from auto-collected data in QUARD-Auto. 
Please note that as an efficient model for data utilization, GeRM's training does not necessitate the use of all the data in the dataset.
This could ensure a fair comparison between GeRM and other imitation learning methods for they shared the fully same successful data. 
To fully harness the learning potential of RL within sub-optimal data, we establish a ratio of demonstration to sub-optimal data at \textbf{75.69\%} and \textbf{24.31\%}, respectively. 
For simplicity, we design sparse rewards: the reward of demonstration is \textbf{1.0}, and sub-optimal data is \textbf{0.0}. More detail can be seen in Figure~\ref{fig:train}.

\textbf{Baseline.}
To evaluate the effectiveness of GeRM and the necessity of the existence of MoE structure and Q-Learning.
We select 2 IL approaches (RT-1\cite{brohan2023rt1}, GeRM w/o RL) and 1 RL approach (GeRM w/o RL) as our baseline. 
Here we adjust RT-1 to suit the quadruped robots.
GeRM w/o RL is our GeRM trained in an imitation learning way instead of RL way and GeRM w/o MoE is GeRM ablating the MoE structure.

\textbf{Evaluation Details.}
We conducted a comprehensive and robust series of experiments. 
To ensure data fidelity and mitigate the impact of stochastic variability, our primary experiments for each model encompassed the entirety of tasks including all \textbf{99} sub-tasks, with \textbf{400} trajectories meticulously tested for each. 
To evaluate \textbf{Q1}, we evaluate GeRM on different settings of gaits, such as ``trotting", ``bounding", ``pronking", and ``pacing", and different object settings, including seen objects that exist in offline datasets and unseen objects that out of the distribution, to test its performance as a generalist model.
In the experiments pertaining to \textbf{Q4}, 400 trajectories were rigorously evaluated per epoch for each model on a single task. 
Additionally, a subset of experiments was allocated for other necessary activities (e.g. computational cost analysis and visualization). 
Furthermore, employing the autonomous data-collection methodology discussed earlier, we systematically gathered all testing data to facilitate the expansion of our dataset.
~\subsection{Experimental Results}

\textbf{Q1\&Q2. GeRM effectively learns from mix-quality data, outperforms other methods, and demonstrates superior capabilities in multi-tasks with MoE Module and Q Learning playing significant roles in GeRM.}
The experimental results in Table~\ref{tab:results} aim to answer Q1\&Q2.
Since there is only a maximum of \textbf{8610} demonstrations of different tasks, we observe from Table~\ref{tab:results} that an IL algorithm like RT-1 and GeRM w/o RL, which also uses a similar Transformer architecture, struggles to obtain a good performance when learning from the limited pool of demonstrations. 
Offline RL method (GeRM w/o MoE), can learn from both demonstrations and failed episodes, and show better performance compared to RT-1.
Indeed, GeRM trained on demonstrations has exhibited a significant performance improvement, thanks to the model architecture of GeRM itself. 
Furthermore, GeRM trained with the inclusion of sub-optimal data has further enhanced its performance across most tasks, particularly achieving substantial improvements in ``\textit{Stop}" tasks. 
GeRM has the highest success rates and outperforms both the behavior cloning baseline (RT-1, GeRM w/o RL) and offline RL baselines (GeRM w/o MoE), exceeding the performance of the best-performing prior method by \textbf{30\%}-\textbf{70\%}. 
This demonstrates that GeRM can effectively improve upon human demonstrations using autonomously collected sub-optimal data.
It also demonstrates the significance of each component design within GeRM.

~\textbf{Q3. MoE Modules balance computational cost and performance by activating part of the parameter when inference.}
We also compare the parameter counts of each model.
GeRM exhibits efficiency in the cost-performance spectrum (see Table~\ref{tab:results}). 
As sparse Mixture-of-Experts models, GeRM w/o RL and GeRM only use \textbf{39.31M} activated parameters for each token, which means it only uses \textbf{1/2} total parameters and \textbf{1/8} FFN layers. 
With slight parameter increases (only \textbf{5.81M}), GeRM is able to outperform RT-1 across all categories.
Moreover, another MoE model GeRM w/o RL performs better than RT-1 across most categories with the same activated parameters.

Note that this analysis focuses on the active parameter count, which is directly proportional to the inference computational cost, but does not consider the hardware utilization and training costs. 
As for device utilization, we note that the MoE layer introduces additional overhead due to the routing mechanism and the increased memory loads when running more than one expert per device. 
They are more suitable for batched workloads where one can reach a good degree of arithmetic intensity.
For training cost, we will discuss it in the next question.

~\textbf{Q4. GeRM exhibits commendable training efficiency.}\label{Q4}
While GeRM could control its computational cost at a relatively rational level, its efficiency in the training stage may raise concerns.
So we perform a comparison experiment between GeRM and other baselines to assess their performance in the ``\textit{Go to the red cube}" task.
To ensure the same input data volume, we only utilize the demonstration data to exclude potential additional data volume (sub-optimal data).
According to Figure~\ref{fig:trend}, under the same number of epochs, GeRM often achieves higher success rates. 
By the \textbf{2}nd epoch, it has already reached a similar level to that of RT-1's \textbf{20}th epoch and essentially converged by the \textbf{7}th epoch. 
Similarly, GeRM w/o MoE, also an offline RL method, converges in approximately \textbf{8} epochs. 
In contrast, Imitation Learning Methods (GeRM w/o RL, RT-1) fail to converge by the \textbf{10}th epoch. 
It is noteworthy that GeRM's performance, even when exclusively trained with demonstrations, remains impressive.
This observation underscores GeRM's proficiency not only in effectively harnessing sub-optimal data but also in leveraging demonstrations with superior efficiency compared to alternative methodologies. 
\textbf{Such findings serve to further substantiate the efficacy of GeRM in optimizing data utilization strategies.}

~\textbf{Q5. GeRM shows emergent skills in dynamic adaptive path planning.}
Through the RL from the large-scale combination of demonstrations and sub-optimal data, GeRM has the potential to autonomously explore unseen skills beyond the demonstrations, known as emergent skills.
Therefore, we aim to evaluate the degree to which such models can show emergent skills.
We demonstrate an example in Figure~\ref{fig:emergent}.
Taking the task ``\textit{Go to the fan and avoid the obstacle}" as an example, in the upper figure, the quadruped robot's vision is limited at the initial position, hampering its ability to determine the direction of movement. To avoid the obstacle it turns to the left randomly. Subsequently, upon encountering the incorrect visual input, the robot executes a substantial reorientation to align with the correct target outside its original field of view. It then proceeds to steer towards the destination, ultimately accomplishing the task. Notably, such trajectories were out-of-distribution of our training dataset. Conversely, the lower figure illustrates a common failure example by IL ways, the robot chooses the false direction and directly reaches the wrong target.
We find that through our exploration GeRM inherits novel capabilities in terms of \textbf{dynamic adaptive path planning} in the context of the scene, which means it can make decisions, plan future paths, and change next-step action according to the visual perception. 

%% file: paper/Conclusion.tex
\section{Conclusion, Limitations and Future Work}

\begin{figure}[t]
  \centering
  \includegraphics[width=0.5\textwidth]{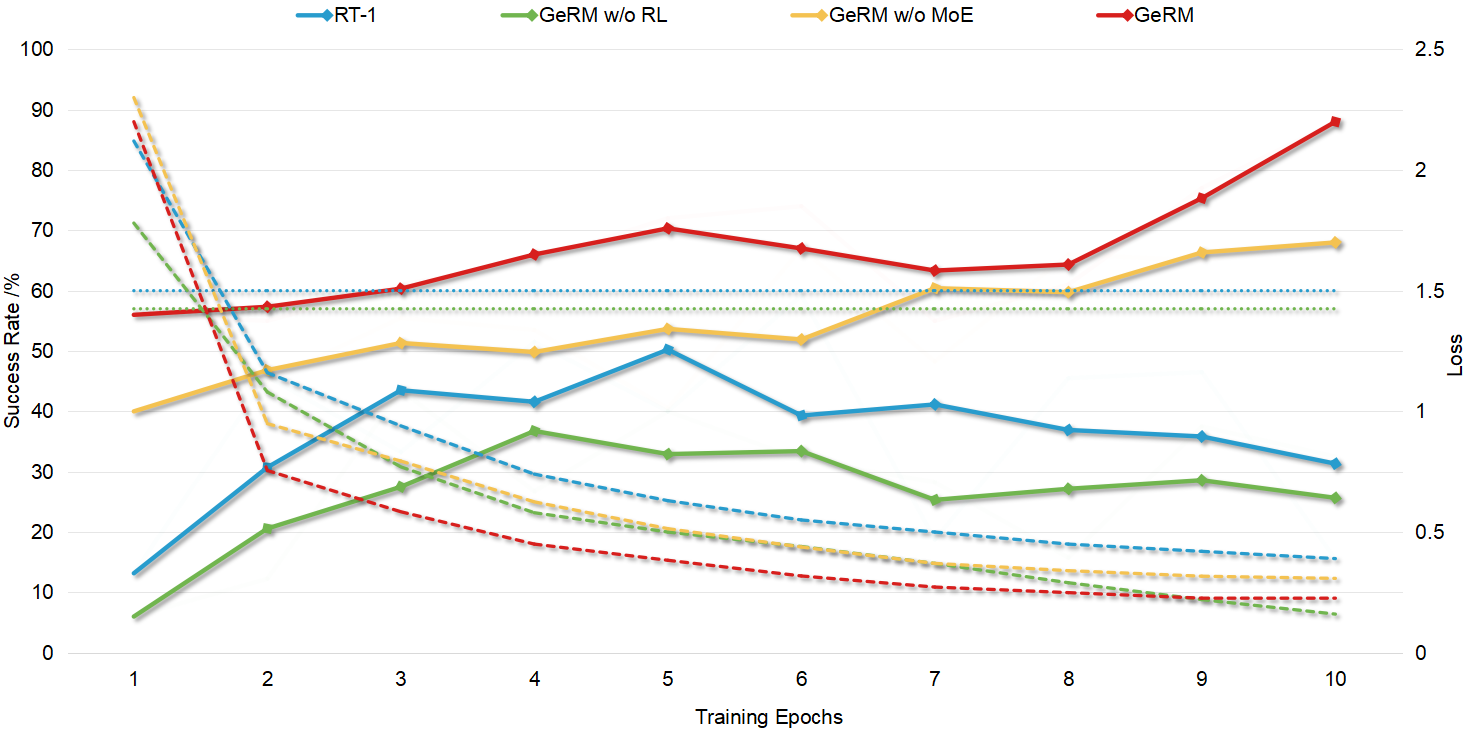}
  \caption{\textbf{Performance change and Loss on ``\textit{Go to the red cube}" task.} Solid lines represent the success rate, dotted lines represent the final success rate for 20 epochs, and dashed lines represent loss. Note: RL approaches employ MSE loss, which should be scaled by 0.1, while IL ways employ Cross-Entropy as the loss function.}
  \label{fig:trend}
  \vspace{-0.1in}
\end{figure}

We have presented GeRM, the first Mixture-of-Experts model for quadruped reinforcement learning.
We have surpassed the limitations of quadruped robots in demonstration by using RL, enhancing the ability and efficiency of data utilization, with the potential to elevate robot performance to super-human levels. 
By incorporating the transformer-based MoE model, we have expanded the model's capacity and reinforced its capabilities, enabling it to possess generalist abilities in multi-task. 
Our model achieves high performance with the limited computational cost, while further optimizing the data utilization capabilities and fostering the development of emergent skills.
We introduce QUARD-Auto, a dataset comprising both successful and failed task data, totaling 257k trajectories, serving as a benchmark for robotic imitation learning and reinforcement learning in the future, which could benefit the robot learning community.

\textbf{Limitations \& Future Work.} 
\wx{1.} While our model demonstrates effectiveness for quadruped robots in simulation, our next step involves extending its capabilities to real-world scenarios. We aim to assess its performance in real-world environments and conduct additional research to ensure its adaptability to real-world settings.
\wx{2.} With aspirations for our model, GeRM, to excel across a broader range of tasks as a generalist, our future endeavors involve expanding its proficiency. To achieve this, we intend to curate a larger dataset encompassing a wider array of task categories. This will enable us to further evaluate the robustness of GeRM and its ability to generalize effectively.